\pdfoutput=1

\documentclass[11pt]{article}

\usepackage[]{EMNLP2023}

\usepackage{times}
\usepackage{latexsym}

\usepackage[T1]{fontenc}

\usepackage[utf8]{inputenc}

\usepackage[verbose]{microtype}

\usepackage{inconsolata}

\usepackage{multirow}
\usepackage{xcolor}

\usepackage{graphicx}
\usepackage{amsmath} 

\usepackage{booktabs} 
\usepackage{siunitx}  

\usepackage{url}

\usepackage{kotex}
 \usepackage[most]{tcolorbox}

\newtcolorbox{promptbox}[2][]{
  breakable,
  title=#2,
  width=\linewidth,
  colback=gray!5,
  colframe=gray!50!black,
  boxrule=0.8pt,
  arc=4pt,
  left=4pt, right=4pt, top=4pt, bottom=4pt,
  fontupper=\small\ttfamily,
  #1
}

\newtcolorbox{infobox}[2][]{
  breakable,
  title=#2,
  width=\linewidth,
  colback=gray!5,
  colframe=gray!50!black,
  boxrule=0.8pt,
  arc=4pt,
  left=4pt, right=4pt, top=4pt, bottom=4pt,
  #1
}

\title{Format Inertia: A Failure Mechanism of LLMs \\ in Medical Pre-Consultation}

\author{
    \textbf{Seungseop Lim$^{1}$, Gibaeg Kim$^{1}$, Wooseok Han$^{1}$}, \\
    \textbf{Jean Seo$^{1}$, Hyunkyung Lee$^{1}$, Jaehyo Yoo$^{1}$, Eunho Yang$^{1,2}$}\thanks{\ \ Corresponding author.} \\
    $^{1}$AITRICS \qquad
    $^{2}$KAIST \\
    \texttt{\{ss.lim, eunhoy\}@aitrics.com, eunhoy@kaist.ac.kr}
}

\begin{document}

\maketitle


\begin{abstract}
Recent advances in Large Language Models (LLMs) have brought significant improvements to various service domains, including chatbots and medical pre-consultation applications. In the healthcare domain, the most common approach for adapting LLMs to multi-turn dialogue generation is Supervised Fine-Tuning (SFT). However, datasets for SFT in tasks like medical pre-consultation typically exhibit a skewed turn-count distribution. Training on such data induces a novel failure mechanism we term \textit{Format Inertia}, where models tend to generate repetitive, format-correct, but diagnostically uninformative questions in long medical dialogues. To mitigate this observed failure mechanism, we adopt a simple, data-centric method that rebalances the turn-count distribution of the training dataset. Experimental results show that our approach substantially alleviates Format Inertia in medical pre-consultation.
\end{abstract}

\section{Introduction} 

The rapid advancement of Large Language Models (LLMs) has revolutionized the field of conversational artificial intelligence, significantly enhancing user experiences across various domains \citep{achiam2023gpt, touvron2023llama}. However, the majority of these successes have predominantly focused on single-turn interactions. In contrast, real-world industrial applications, particularly in healthcare domains such as medical pre-consultation, demand robust multi-turn dialogues between patients and doctors \cite{tu2024towards, saab2025advancing, hu2024uncertainty, winston2024medical}. These complex multi-turn environments require the model to effectively maintain and utilize long-range conversational context, which remains a challenging problem \cite{laban2025llms}.

\begin{figure*}[t]
    \centering
    \includegraphics[width=1.0\textwidth]{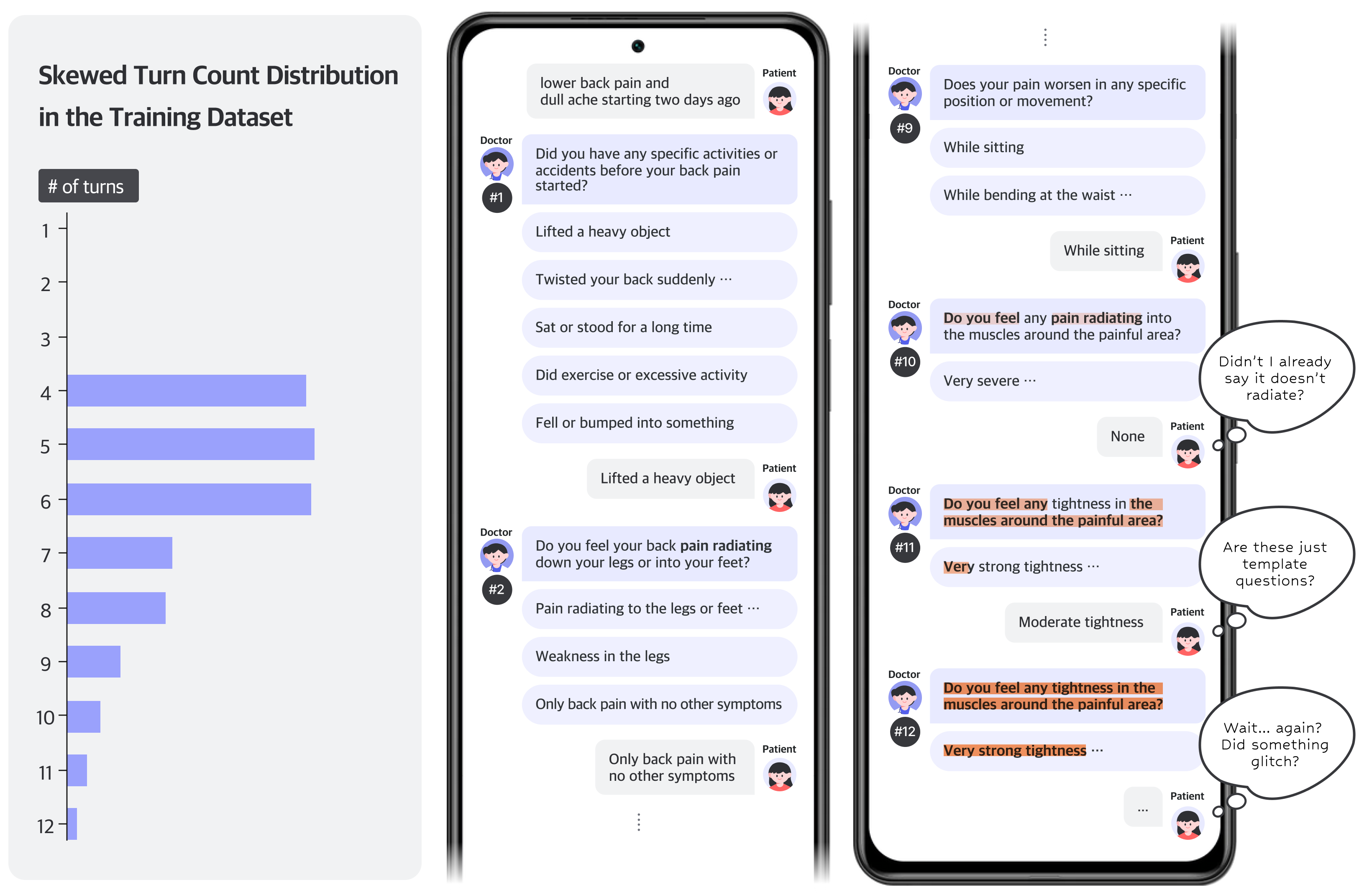}
    \caption{Example of \textbf{Format Inertia} in Medical Pre-Consultation. When trained on skewed turn-count distribution, the model overly relies on previously generated question patterns—preserving superficial format but failing to contribute new diagnostic information (\#2→\#10) and repeating identical questions (\#11→\#12). 
    Format Inertia not only stalls clinical progress but also leaves the patient feeling confused, thereby undermining the overall user experience.
    }
    \label{fig:fig1}
\end{figure*}

The predominant approach to adapting LLMs for multi-turn dialogue generation in healthcare domains is Supervised Fine-Tuning (SFT)~\citep{wang2023customizing, yu2025finemedlm, christophe2024beyond}. Despite its widespread use, prior work has largely overlooked the effects of skewed turn-count distribution in training data, particularly their influence on an LLM’s ability to maintain long-range conversational context and the specific failure mechanisms that emerge in long medical dialogues. While fine-tuning on high-quality medical datasets is widely recognized as critical for improving accuracy, consistency, and reliability, the role of turn-count distribution in training data in shaping long medical dialogue competence remains insufficiently understood.

This paper aims to fill this gap by hypothesizing and empirically validating that the skewed turn-count distribution in the training datasets impairs LLMs’ adherence to task constraints in medical pre-consultation multi-turn dialogues. Specifically, models trained predominantly on dialogues with short turn lengths—reflecting real-world production data distribution—lack sufficient exposure to the complex contextual dependencies present in longer interviews. Consequently, these models struggle to generate follow-up questions that acquire novel diagnostic information during long medical dialogues.

To better understand this failure, we introduce a novel failure mechanism termed \emph{Format Inertia}, which we define and empirically observe in this study. Analogous to physical inertia, where an object maintains its current state absent external force, Format Inertia describes LLMs' tendency to overly rely on previously generated question patterns when confronted with uncertainty in rarely seen long medical dialogues. This results in repetitive and unproductive questioning that preserves superficial format but fails to contribute new diagnostic information, as illustrated in Figure \ref{fig:fig1}, where models trained on skewed turn-count distribution repeat previous questions in the latter turns of long dialogues.

To address this issue, we adopt a data-centric strategy that constructs a Uniform Turn-Count Dataset by sampling an equal number of dialogues across max turn-count bins, thereby ensuring balanced exposure to diverse dialogue lengths. This approach significantly mitigates Format Inertia in long medical pre-consultations. Our main contributions are summarized as follows:
\begin{itemize}
    \item \textbf{Analysis of Turn-Count Distribution Impact:} We systematically identify and quantitatively analyze the causal relationship between skewed turn-count distribution in the training dataset and degraded task constraint adherence in LLMs. This clarifies a core reason for performance degradation when training medical pre-consultation models under realistic conditions.
    
    \item \textbf{Definition of Format Inertia:} We propose and empirically observe \emph{Format Inertia}, a novel phenomenon where models trained on skewed turn-count distribution generate repetitive and unproductive medical questions by over-relying on existing question formats when faced with uncertainty in long medical dialogues. To our knowledge, this study is the first to specifically explain the failure of LLMs to maintain context when trained with a skewed turn-count distribution.
    
    \item \textbf{Practical Data-Centric Solution:} We adopt a data-centric approach that rebalances the turn-count distribution in the training dataset by constructing a Uniform Turn-Count Dataset. This simple adjustment significantly improves the model’s ability to maintain long-range conversational context and adhere to task constraints in long medical dialogues. 
\end{itemize}

\section{Background}
\paragraph{Challenges in Multi-turn Conversational AI}
Multi-turn dialogue systems must maintain context across long dialogues while satisfying various constraints. Recent benchmarks evaluate these capabilities: FollowBench~\cite{jiang2023followbench} assesses format and content constraints, CFBench~\cite{zhang2024cfbench} provides a comprehensive evaluation, and \citet{wen2024benchmarking} considers constraint combinations. These studies highlight the complexity of real-world instruction-following requirements.

To address these challenges, methodologies like Parrot~\cite{sun2024parrot} enhance multi-turn capabilities, while \citet{ren2025step} focus on soft constraints requiring contextual judgment. However, existing research primarily concentrates on model architectures or training paradigms, overlooking how statistical characteristics of training data affect long dialogue performance.

\paragraph{Conversational Information Seeking in Medical Domain}
Medical pre-consultation systematically collects patients' History of Present Illness (HPI) for diagnostic decisions. Systems must actively seek diagnostically valuable information beyond scripted questions~\cite{tu2024towards, saab2025advancing}. Evaluation frameworks like CRAFT-MD~\cite{johri2025evaluation} and planning methods like Uncertainty of Thoughts~\cite{hu2024uncertainty} support meaningful information seeking in uncertain medical contexts.

While Supervised Fine-Tuning (SFT) using online medical consultation records or synthetic medical data has achieved high performance~\cite{wang2023customizing, yu2025finemedlm}, real medical data naturally exhibits a distribution skewed toward shorter dialogues. The impact of such skewed turn-count distribution on long dialogue performance remains unexplored.

\section{Methodology}
In this section, we investigate how the turn-count distribution in the training dataset affects Large Language Models (LLMs) engaged in medical pre-consultation. We first formalize the task, then describe a failure mechanism triggered by skewed turn-count distribution, and finally present a data-centric mitigation strategy.

\subsection{Task Definition}
The objective of this task is to generate contextually appropriate questions in a multi-turn dialogue setting, specifically within the domain of medical pre-consultation scenarios between patients and doctors. In this scenario, the LLM acts as a doctor who iteratively interacts with a patient. Beyond producing formally correct questions, the model must maintain and use the accumulated conversational history to ask contextually appropriate follow-up questions that diagnostically informative.

These requirements can be grouped into two sets of constraints: (i) format constraints, which govern structural aspects such as response format, response language, and the absence of forbidden terms; and (ii) task constraints, which ensure that each generated question is clinically meaningful and advances the diagnostic goal.

\subsection{Format Inertia: A Failure Mechanism Induced by Skewed Turn-Count Distribution}
As in most real-world datasets, medical dialogues exhibit a skewed turn-count distribution: short conversations dominate, while long medical dialogues are scarce. When an LLM is fine-tuned on such data, its exposure to long dialogues is limited. We hypothesise that this imbalance leads to a failure mechanism we call \emph{Format Inertia}. Faced with these under-represented cases, the model tends to safe, repetitive question templates that satisfy superficial format constraints but fail to acquire new diagnostic information.

Format Inertia manifests as an over-reliance on previously generated patterns. The model preserves surface form—e.g., a fixed question pattern—yet produces redundant or low-utility questions because it cannot integrate long-range context effectively. We validate this hypothesis empirically in Section~\ref{ssec:analysis}.

\subsection{Uniform Turn-Count Dataset: Mitigating Format Inertia}

To counteract Format Inertia, we adopt a simple data-centric remedy: constructing a Uniform Turn-Count Dataset. This approach rebalances the training data by ensuring equal representation for dialogues of varying lengths. By exposing the model to a balanced mix of short and long conversations, we mitigate its tendency to follow repetitive patterns when faced with less familiar, long dialogues. The dataset construction process is as follows:

\begin{enumerate}
    \item \textbf{Turn-Based Binning:} All $N$ dialogues from the source dataset are grouped into $B$ bins based on their maximum turn-count.
    
    \item \textbf{Quota Determination:} A sampling quota, $q$, is set to the number of dialogues in the smallest bin.
    
    \item \textbf{Uniform Sampling:} We select all bins that meet a minimum turn threshold, $T_{\text{min}}$. Let $B'$ be the number of selected bins. From each of these $B'$ bins, we randomly sample $q$ dialogues.
    \item \textbf{Dataset Assembly:} The sampled dialogues are aggregated to form the final Uniform Turn-Count Dataset, containing a total of $q \times B'$ dialogues.
\end{enumerate}

Equalizing turn-counts naturally broadens the spectrum of clinical scenarios encountered during training. Shorter dialogues often correspond to consultations with patients having relatively minor conditions, while longer dialogues are typically associated with patients who require more in-depth medical reasoning, involving complex history-taking. This balanced exposure ensures the model develops robust strategies for a wide range of consultation lengths, enhancing its ability to handle the diverse patient interactions found in real-world clinical settings.

\begin{table*}[t]
\centering
\resizebox{0.9\textwidth}{!}{%


\begin{tabular}{@{}l p{\textwidth}@{}}
\toprule
\textbf{Constraint} & \textbf{Description} \\
\midrule
\multicolumn{2}{@{}l}{\textbf{Format Constraints}} \\
\cmidrule(r){1-2}
\texttt{response\_format} & Output adheres to the required format. \\
\texttt{response\_language} & Response is consistently written in the designated language. \\
\texttt{forbidden\_words} & Prohibited terms are entirely excluded from the response. \\
\texttt{number\_options} & Presents the correct number of options as specified. \\
\texttt{sentence\_startend} & Each sentence begins and ends with the appropriate grammatical form, such as polite endings. \\
\midrule
\multicolumn{2}{@{}l}{\textbf{Task Constraints}} \\
\cmidrule(r){1-2}
\texttt{clinical\_utility} & The response effectively elicits additional clinical information relevant to diagnosis. \\
\bottomrule
\end{tabular}
}
\caption{Definitions of Format and Task Constraints used for model evaluation. Specific examples for each constraint category are provided in Appendix~\ref{appendix:constraint_examples}.}
\label{tab:table1}
\end{table*}

\section{Experiments}
In this section, we present experiments evaluating our central hypothesis and method in our medical pre-consultation service scenarios.

\subsection{Experimental Setting}
\label{sec:experimental_setting}

\paragraph{Real-World Data Source} 
To test our hypothesis, we gathered a corpus of approximately 8,000 medical pre-consultation dialogues generated by 40 doctors interacting with a patient model. This dataset naturally exhibits a skewed turn-count distribution, with a majority of short conversations and a minority of long ones, reflecting typical real-world scenarios. Further details on data collection are available in Appendix~\ref{appendix:data}.

\paragraph{Training Datasets} \label{training_dataset} To systematically analyze the effect of turn-count distribution, we constructed three training datasets sampled from the real-world corpus: (i) a Skewed Turn 1k subset (1,000 dialogues), (ii) the full Skewed Turn 8k set (8,000 dialogues), and (iii) a Uniform Turn 1k subset (1,000 dialogues). 

\begin{itemize}
    \item \textbf{Skewed Turn-Count Dataset:} This dataset mirrors the natural skewed turn-count distribution of the real-world data source, and we evaluate it at two different scales (1,000 and 8,000) to investigate the effect of data volume under the same distribution.
    \item \textbf{Uniform Turn-Count Dataset:} This dataset is constructed using our data-centric strategy, which alleviates distributional bias by uniformly sampling dialogues across the entire spectrum of turn lengths observed in the real-world dataset. This approach ensures balanced exposure to both short and long interactions during training.
\end{itemize}

\paragraph{Evaluation Datasets} 
To enable consistent and rigorous evaluation of long-form dialogue capabilities, we constructed a dedicated evaluation set. This set comprises 100 patient profiles curated by medical professionals. During evaluation, each model engages in a simulated multi-turn dialogue with a patient model instantiated from these profiles.

\paragraph{Doctor Model} 
Our experiments included both open-source and proprietary models. For the open-source group, we selected Gemma-3-4B\cite{team2025gemma} and Qwen2.5-3B\cite{yang2025qwen3}, based on their strong performance and reliable multilingual language support, including Korean, which is the target service language.

As a representative proprietary model, we employed GPT-4.1-mini\footnote{gpt-4.1-mini-2025-04-14}, chosen for its balance between state-of-the-art capability and computational efficiency.
All doctor models were provided with a simplified version of the patient's condition to simulate limited prior knowledge typically available during medical pre-consultation. (see Appendix~\ref{appendix:patient_info} for details).

\paragraph{Patient Model}
To ensure consistent and reproducible evaluation across different doctor models, we used o4-mini\footnote{o4-mini-2025-04-16} as the patient model throughout all inference-time interactions.
The patient model was equipped with full knowledge of each clinical case, including detailed medical history and contextual information. This setup emulates a realistic pre-consultation scenario, where the patient provides accurate and consistent responses, thereby allowing fair comparisons across doctor model outputs.

\begin{table}[t]
\centering
\resizebox{0.8\columnwidth}{!}{%

\begin{tabular}{llccc}
    \toprule
    \textbf{Model} & \textbf{Type} & \textbf{Samples} & \textbf{FCSR} & \textbf{TCSR} \\
    \midrule
    \multirow{4}{*}{Gemma-3 (4B)} 
        & base & - & 0.361 & 0.872 \\
        \cmidrule(lr){2-5}
        & skew & 1k & 0.960 & 0.824 \\
        & skew & 8k & 0.966 & 0.811 \\
        & uniform & 1k & \textbf{0.967} & \textbf{0.891} \\
    \midrule
    \multirow{4}{*}{Qwen2.5 (3B)} 
        & base & - & 0.363 & 0.783 \\
        \cmidrule(lr){2-5}
        & skew & 1k & 0.922 & 0.746 \\
        & skew & 8k & 0.914 & 0.737 \\
        & uniform & 1k & \textbf{0.927} & \textbf{0.812} \\
    \midrule
    GPT-4.1-mini & base & - & 0.906 & 0.880 \\
    \bottomrule
\end{tabular} %
}
\caption{Comparison of Format-Constraint Satisfaction Rate (FCSR) and Task-Constraint Satisfaction Rate (TCSR) across models trained on datasets with varying turn-count distribution. The `base' denotes the original model before fine-tuning. ‘Samples’ refers to the number of dialogues used for fine-tuning.}
\label{tab:table2}
\end{table}

\paragraph{Evaluation Metrics}
To evaluate the output quality of LLMs in the context of medical pre-consultation under production settings, we assess two critical aspects: adherence to service-specific format constraints and the clinical utility of the generated questions.
To measure adherence to structural requirements, we use the \emph{Format-Constraint Satisfaction Rate (FCSR)}. This metric evaluates the model’s compliance with five predefined formatting rule categories, based on a modified version of the benchmark proposed in~\cite{zhou2023instruction}, as listed in Table~\ref{tab:table1}. We implement a strictly coded validator to automatically and consistently check each dialogue turn against all constraint categories.
FCSR is computed as the proportion of dialogue turns that fully satisfy all format constraints out of the total number of evaluated turns:
\[
\scalebox{1.0}{$
\text{FCSR} = \frac{\text{\# of Turns Satisfying All Format Constraints}}{\text{\# of Turns in Total}}
$}
\]

To assess the contextual appropriateness and diagnostic utility of the model’s outputs, we define the \emph{Task-Constraint Satisfaction Rate (TCSR)}. This metric evaluates whether the model adheres to qualitative task constraints that ensure the generation of clinically meaningful questions, as detailed in Table~\ref{tab:table1}.
TCSR is computed as the fraction of turns that fulfill task-specific criteria:
\[
\scalebox{1.0}{$
\text{TCSR} = \frac{\text{\# of Turns Satisfying All Task Constraints}}{\text{\# of Turns in Total}}
$}
\]

To verify the diagnostic utility of the generated questions, we adopt an LLM-as-a-Judge approach~\cite{zheng2023judging} as our evaluation framework (see Appendix~\ref{appendix:evaluation} for details).

To ensure the reliability of our evaluation framework, we conducted a human verification study on a subset of 240 samples from the evaluation data. Two medical experts independently assessed these samples using the same criteria as our LLM judge. The inter-rater agreement analysis revealed:
\begin{itemize}
    \item \textbf{Cohen's Kappa:} 0.8091, indicating ``almost perfect'' agreement according to Landis and Koch's interpretation scale.
    \item \textbf{Spearman's Rank Correlation:} 0.8129 (p < 0.0001), confirming a strong positive correlation between human and LLM assessments.
\end{itemize}
This high level of agreement validates that our LLM-based evaluation provides reliable and scalable assessment for this specific, constrained task.

\subsection{Results}
\label{ssec:results}
Our experiments reveal that while SFT improves format adherence, the turn-count distribution of the training data critically impacts task performance. 
As shown in Table~\ref{tab:table2}, models fine-tuned on the skewed turn-count dataset with 1,000 samples achieve a high FCSR but suffer a TCSR drop—for example, from 0.872 to 0.824 in Gemma-3 and from 0.783 to 0.746 in Qwen2.5—demonstrating Format Inertia. Furthermore, increasing the volume of skewed data to 8,000 samples can exacerbate this degradation, leading to an even lower TCSR.

In contrast, fine-tuning on the Uniform turn-count dataset with 1,000 samples effectively mitigates this trade-off. Specifically, uniform sampling recovers TCSR to 0.891 while slightly improving FCSR to 0.967 for Gemma-3, enabling the model to generate responses that are both formally correct and clinically meaningful. This finding underscores that for medical pre-consultation tasks, data quality and distributional balance are paramount over sheer quantity, as the model trained on 1,000 uniform samples significantly outperforms the one trained on 8,000 skewed samples.

\begin{figure}[t]
    \centering
\includegraphics[width=0.9\columnwidth]{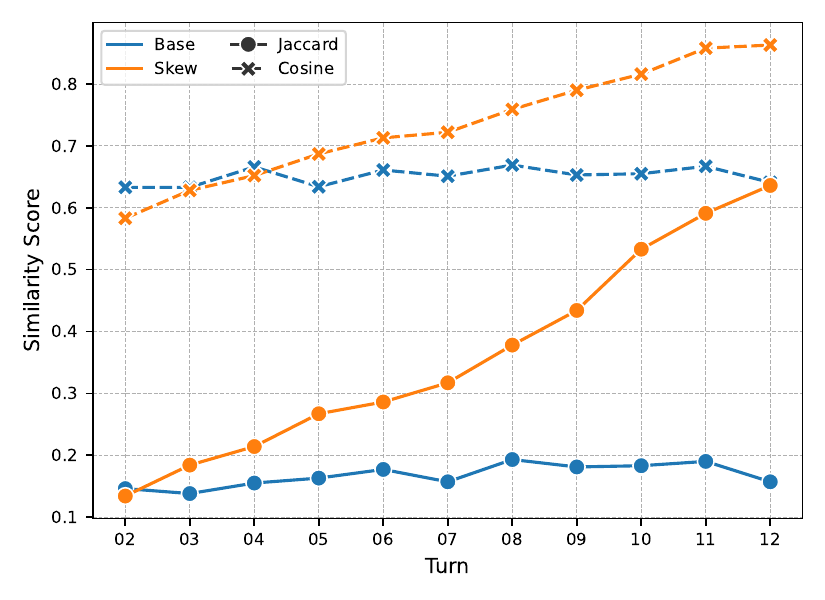}
    \caption{Models trained on skewed turn-count data show a progressive increase in Jaccard and Cosine similarity across dialogue turns, indicating an intensifying pattern of repetitive questioning driven by Format Inertia, in contrast to the base model.}
    \label{fig:fig2}
\end{figure}

\begin{figure}[t]
    \centering
\includegraphics[width=0.9\columnwidth]{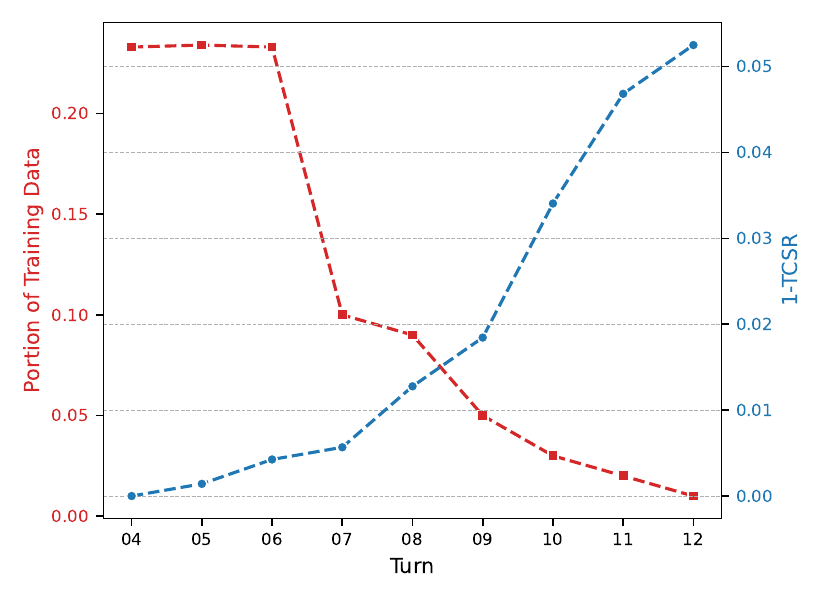}
    \caption{Impact of Skewed turn-count data on TCSR. Inverse relationship between the frequency of turns in the Skewed Turn training data (left y-axis) and the Task-Constraint failure rate (1-TCSR) in evaluation (right y-axis), highlighting performance degradation on underrepresented long turns.}
    \label{fig:fig3}
\end{figure}

\begin{table}[t]
\centering
\resizebox{0.8\columnwidth}{!}{%



\begin{tabular}{lllcc}
\toprule
\textbf{Model} & \textbf{Size} & \textbf{Type} & \textbf{FCSR} & \textbf{TCSR} \\
\midrule
\multirow{9}{*}{Gemma-3} & \multirow{3}{*}{4B} & base & 0.361 & 0.872 \\
\cmidrule(lr){3-5}
 & & skew & 0.960 & 0.824 \\
 & & uniform & \textbf{0.967} & \textbf{0.891} \\
\cmidrule(lr){2-5} 
 & \multirow{3}{*}{12B} & base & 0.577 & 0.880 \\
 \cmidrule(lr){3-5}

 & & skew & 0.972 & 0.843 \\
 & & uniform & \textbf{0.976} & \textbf{0.896} \\
\cmidrule(lr){2-5} 
 & \multirow{3}{*}{27B} & base & 0.863 & 0.884 \\
 \cmidrule(lr){3-5}
 & & skew & 0.963 & 0.853 \\
 & & uniform & \textbf{0.964} & \textbf{0.901} \\
\bottomrule
\end{tabular}

}
\caption{Performance variation by model size, illustrating the impact of Uniform Turn-Count Dataset on the Gemma-3 model across its different scales.}
\label{tab:table3}
\end{table}

\subsection{Analysis}
\label{ssec:analysis}
Our analysis identifies the root cause of performance degradation as \emph{Format Inertia}. This phenomenon describes the model's tendency to follow familiar, repetitive questioning patterns when faced with the uncertainty of long medical dialogue contexts, which are under-represented in skewed training data. In such cases, the model successfully maintains surface-level format constraints but neglects the more cognitively demanding task of generating novel, contextually relevant questions.

We quantify this failure mechanism by analyzing question similarity throughout the dialogue. To probe this, we measure both lexical and semantic similarity of model-generated questions during medical pre-consultation. Specifically, we use Jaccard Similarity for lexical comparison and Cosine Similarity for semantic similarity, computing each question’s maximum similarity against all preceding ones. As illustrated in Figure~\ref{fig:fig2}, models trained on skewed turn-count data show a progressive increase in both metrics, confirming that they produce increasingly redundant questions in both form and meaning. In contrast, the base model without fine-tuning shows no such trend. This behavior strongly indicates that \emph{Format Inertia} is a direct consequence of the training data's skewed dialogue structure, causing the model to repeat surface-level patterns that are grammatically sound but semantically unproductive. Figure~\ref{fig:fig3} demonstrates a clear inverse relationship: a given turn length's rarity in the training data correlates with a higher task-constraint failure rate (1-TCSR) during evaluation. This confirms that Format Inertia is a direct consequence of the model's limited exposure to long medical dialogues, forcing it to fall back on formally correct but functionally redundant patterns.

The issue is particularly critical for efficiency-oriented models commonly used in real-world deployments. As shown in Table~\ref{tab:table3}, although the benefits of a uniform turn-count distribution are observed across all model sizes, TCSR consistently declines when models are trained on skewed data but recovers under the uniform setting. These results confirm that Format Inertia is a persistent challenge regardless of scale, underscoring the importance of addressing it to ensure the reliability and practicality of conversational AI systems.

\section{Conclusion}
In this study, we identify Format Inertia as a failure mechanism in LLM-based medical pre-consultation, caused by skewed turn-count distribution in the training data. Limited exposure to long dialogues leads models to tend to familiar question patterns, resulting in repetitive, low-utility outputs that meet format requirements but fail to elicit new diagnostic information. To address this, we construct a Uniform Turn-Count Dataset, which rebalances turn distribution. Our experiments show this approach effectively mitigates Format Inertia in long medical dialogues. These findings underscore the critical role of data distribution, especially the turn-count distribution, in multi-turn robustness.

\section*{Limitations}
Our study has several limitations that also point to directions for future work. 
First, our experiments are concentrated on the specific domain of medical pre-consultation. A valuable direction for future research would be to investigate whether the \emph{Format Inertia} phenomenon we identified generalizes to other conversational domains, such as legal counseling or technical support, which would broaden the implications of our findings.
Second, our evaluation of question quality relied on an LLM-as-a-Judge approach. While this method ensures consistency, it may not fully capture the nuanced judgments of human medical experts. Future work could strengthen the validity of our evaluation framework by systematically analyzing the correlation between LLM-based assessments and human expert judgments.
Finally, while we identify the skewed turn-count distribution as a key driver of Format Inertia, other factors—such as the diversity of dialogue scenarios or specific model architectures—could also contribute to this phenomenon. Exploring these additional factors represents an important avenue for developing a more comprehensive understanding of this failure mode.

\section*{Ethics Statement}
This study was conducted with a strong commitment to ethical principles. The dataset used for training and evaluation was sourced from a controlled patient simulation platform, where medical professionals generated dialogue data. Crucially, this process did not involve any real patient data, thereby avoiding risks related to patient privacy and data confidentiality. All data was carefully processed to ensure no personally identifiable information was included. We emphasize that the models developed in this research are intended as assistive tools for pre-consultation and are not a substitute for professional medical diagnosis or advice. Any real-world deployment of such technology would require rigorous safety testing, regulatory approval, and a human-in-the-loop system to ensure patient safety and well-being.


\bibliography{anthology,custom}

\begin{thebibliography}{22}
\expandafter\ifx\csname natexlab\endcsname\relax\def\natexlab#1{#1}\fi

\bibitem[{Achiam et~al.(2023)Achiam, Adler, Agarwal, Ahmad, Akkaya, Aleman, Almeida, Altenschmidt, Altman, Anadkat et~al.}]{achiam2023gpt}
Josh Achiam, Steven Adler, Sandhini Agarwal, Lama Ahmad, Ilge Akkaya, Florencia~Leoni Aleman, Diogo Almeida, Janko Altenschmidt, Sam Altman, Shyamal Anadkat, et~al. 2023.
\newblock Gpt-4 technical report.
\newblock \emph{arXiv preprint arXiv:2303.08774}.

\bibitem[{Christophe et~al.(2024)Christophe, Raha, Maslenkova, Salman, Kanithi, Pimentel, and Khan}]{christophe2024beyond}
Cl{\'e}ment Christophe, Tathagata Raha, Svetlana Maslenkova, Muhammad~Umar Salman, Praveen~K Kanithi, Marco~AF Pimentel, and Shadab Khan. 2024.
\newblock Beyond fine-tuning: Unleashing the potential of continuous pretraining for clinical llms.
\newblock \emph{arXiv preprint arXiv:2409.14988}.

\bibitem[{Dettmers et~al.(2023)Dettmers, Pagnoni, Holtzman, and Zettlemoyer}]{dettmers2023qlora}
Tim Dettmers, Artidoro Pagnoni, Ari Holtzman, and Luke Zettlemoyer. 2023.
\newblock Qlora: Efficient finetuning of quantized llms.
\newblock \emph{Advances in neural information processing systems}, 36:10088--10115.

\bibitem[{Hu et~al.(2024)Hu, Liu, Feng, Zhao, Ng, Luu, He, Koh, and Hooi}]{hu2024uncertainty}
Zhiyuan Hu, Chumin Liu, Xidong Feng, Yilun Zhao, See-Kiong Ng, Anh~Tuan Luu, Junxian He, Pang Wei~W Koh, and Bryan Hooi. 2024.
\newblock Uncertainty of thoughts: Uncertainty-aware planning enhances information seeking in llms.
\newblock \emph{Advances in Neural Information Processing Systems}, 37:24181--24215.

\bibitem[{Jiang et~al.(2023)Jiang, Wang, Zeng, Zhong, Li, Mi, Shang, Jiang, Liu, and Wang}]{jiang2023followbench}
Yuxin Jiang, Yufei Wang, Xingshan Zeng, Wanjun Zhong, Liangyou Li, Fei Mi, Lifeng Shang, Xin Jiang, Qun Liu, and Wei Wang. 2023.
\newblock Followbench: A multi-level fine-grained constraints following benchmark for large language models.
\newblock \emph{arXiv preprint arXiv:2310.20410}.

\bibitem[{Johri et~al.(2025)Johri, Jeong, Tran, Schlessinger, Wongvibulsin, Barnes, Zhou, Cai, Van~Allen, Kim et~al.}]{johri2025evaluation}
Shreya Johri, Jaehwan Jeong, Benjamin~A Tran, Daniel~I Schlessinger, Shannon Wongvibulsin, Leandra~A Barnes, Hong-Yu Zhou, Zhuo~Ran Cai, Eliezer~M Van~Allen, David Kim, et~al. 2025.
\newblock An evaluation framework for clinical use of large language models in patient interaction tasks.
\newblock \emph{Nature medicine}, 31(1):77--86.

\bibitem[{Kweon et~al.(2024)Kweon, Choi, Chu, Song, Hyeon, Gan, Kim, Kim, Park, and Choi}]{kweon2024kormedmcqa}
Sunjun Kweon, Byungjin Choi, Gyouk Chu, Junyeong Song, Daeun Hyeon, Sujin Gan, Jueon Kim, Minkyu Kim, Rae~Woong Park, and Edward Choi. 2024.
\newblock Kormedmcqa: Multi-choice question answering benchmark for korean healthcare professional licensing examinations.
\newblock \emph{arXiv preprint arXiv:2403.01469}.

\bibitem[{Laban et~al.(2025)Laban, Hayashi, Zhou, and Neville}]{laban2025llms}
Philippe Laban, Hiroaki Hayashi, Yingbo Zhou, and Jennifer Neville. 2025.
\newblock Llms get lost in multi-turn conversation.
\newblock \emph{arXiv preprint arXiv:2505.06120}.

\bibitem[{Ren et~al.(2025)Ren, Zeng, He, Liang, Xiao, Zhou, Sun, and Yu}]{ren2025step}
Qingyu Ren, Jie Zeng, Qianyu He, Jiaqing Liang, Yanghua Xiao, Weikang Zhou, Zeye Sun, and Fei Yu. 2025.
\newblock Step-by-step mastery: Enhancing soft constraint following ability of large language models.
\newblock \emph{arXiv preprint arXiv:2501.04945}.

\bibitem[{Saab et~al.(2025)Saab, Freyberg, Park, Strother, Cheng, Weng, Barrett, Stutz, Tomasev, Palepu et~al.}]{saab2025advancing}
Khaled Saab, Jan Freyberg, Chunjong Park, Tim Strother, Yong Cheng, Wei-Hung Weng, David~GT Barrett, David Stutz, Nenad Tomasev, Anil Palepu, et~al. 2025.
\newblock Advancing conversational diagnostic ai with multimodal reasoning.
\newblock \emph{arXiv preprint arXiv:2505.04653}.

\bibitem[{Sun et~al.(2024)Sun, Liu, Zhou, Huang, Song, Zhao, Zhang, Zhang, and Gai}]{sun2024parrot}
Yuchong Sun, Che Liu, Kun Zhou, Jinwen Huang, Ruihua Song, Wayne~Xin Zhao, Fuzheng Zhang, Di~Zhang, and Kun Gai. 2024.
\newblock Parrot: Enhancing multi-turn instruction following for large language models.
\newblock In \emph{Proceedings of the 62nd Annual Meeting of the Association for Computational Linguistics (Volume 1: Long Papers)}, pages 9729--9750.

\bibitem[{Team et~al.(2025)Team, Kamath, Ferret, Pathak, Vieillard, Merhej, Perrin, Matejovicova, Ram{\'e}, Rivi{\`e}re et~al.}]{team2025gemma}
Gemma Team, Aishwarya Kamath, Johan Ferret, Shreya Pathak, Nino Vieillard, Ramona Merhej, Sarah Perrin, Tatiana Matejovicova, Alexandre Ram{\'e}, Morgane Rivi{\`e}re, et~al. 2025.
\newblock Gemma 3 technical report.
\newblock \emph{arXiv preprint arXiv:2503.19786}.

\bibitem[{Touvron et~al.(2023)Touvron, Lavril, Izacard, Martinet, Lachaux, Lacroix, Rozi{\`e}re, Goyal, Hambro, Azhar et~al.}]{touvron2023llama}
Hugo Touvron, Thibaut Lavril, Gautier Izacard, Xavier Martinet, Marie-Anne Lachaux, Timoth{\'e}e Lacroix, Baptiste Rozi{\`e}re, Naman Goyal, Eric Hambro, Faisal Azhar, et~al. 2023.
\newblock Llama: Open and efficient foundation language models.
\newblock \emph{arXiv preprint arXiv:2302.13971}.

\bibitem[{Tu et~al.(2024)Tu, Palepu, Schaekermann, Saab, Freyberg, Tanno, Wang, Li, Amin, Tomasev et~al.}]{tu2024towards}
Tao Tu, Anil Palepu, Mike Schaekermann, Khaled Saab, Jan Freyberg, Ryutaro Tanno, Amy Wang, Brenna Li, Mohamed Amin, Nenad Tomasev, et~al. 2024.
\newblock Towards conversational diagnostic ai.
\newblock \emph{arXiv preprint arXiv:2401.05654}.

\bibitem[{Wang et~al.(2023)Wang, Zhao, and Sun}]{wang2023customizing}
Wen Wang, Zhenyue Zhao, and Tianshu Sun. 2023.
\newblock Customizing large language models for business context: framework and experiments.
\newblock \emph{arXiv preprint arXiv:2312.10225}.

\bibitem[{Wen et~al.(2024)Wen, Ke, Gu, Wu, Huang, Zhou, Li, Hu, Gao, Xu et~al.}]{wen2024benchmarking}
Bosi Wen, Pei Ke, Xiaotao Gu, Lindong Wu, Hao Huang, Jinfeng Zhou, Wenchuang Li, Binxin Hu, Wendy Gao, Jiaxing Xu, et~al. 2024.
\newblock Benchmarking complex instruction-following with multiple constraints composition.
\newblock \emph{Advances in Neural Information Processing Systems}, 37:137610--137645.

\bibitem[{Winston et~al.(2024)Winston, Winston, Winston, and Winston}]{winston2024medical}
Caleb Winston, Cleah Winston, Claris Winston, and Chloe Winston. 2024.
\newblock \href {https://openreview.net/forum?id=du26Irf5kE} {Medical question-generation for pre-consultation with {LLM} in-context learning}.
\newblock In \emph{GenAI for Health: Potential, Trust and Policy Compliance}.

\bibitem[{Yang et~al.(2025)Yang, Li, Yang, Zhang, Hui, Zheng, Yu, Gao, Huang, Lv et~al.}]{yang2025qwen3}
An~Yang, Anfeng Li, Baosong Yang, Beichen Zhang, Binyuan Hui, Bo~Zheng, Bowen Yu, Chang Gao, Chengen Huang, Chenxu Lv, et~al. 2025.
\newblock Qwen3 technical report.
\newblock \emph{arXiv preprint arXiv:2505.09388}.

\bibitem[{Yu et~al.(2025)Yu, Cheng, Cheng, and Feng}]{yu2025finemedlm}
Hongzhou Yu, Tianhao Cheng, Ying Cheng, and Rui Feng. 2025.
\newblock Finemedlm-o1: Enhancing the medical reasoning ability of llm from supervised fine-tuning to test-time training.
\newblock \emph{arXiv preprint arXiv:2501.09213}.

\bibitem[{Zhang et~al.(2024)Zhang, Zhu, Shen, Luo, Zhang, Liang, Yang, Lin, Qiao, Chen et~al.}]{zhang2024cfbench}
Tao Zhang, Chenglin Zhu, Yanjun Shen, Wenjing Luo, Yan Zhang, Hao Liang, Fan Yang, Mingan Lin, Yujing Qiao, Weipeng Chen, et~al. 2024.
\newblock Cfbench: A comprehensive constraints-following benchmark for llms.
\newblock \emph{arXiv preprint arXiv:2408.01122}.

\bibitem[{Zheng et~al.(2023)Zheng, Chiang, Sheng, Zhuang, Wu, Zhuang, Lin, Li, Li, Xing et~al.}]{zheng2023judging}
Lianmin Zheng, Wei-Lin Chiang, Ying Sheng, Siyuan Zhuang, Zhanghao Wu, Yonghao Zhuang, Zi~Lin, Zhuohan Li, Dacheng Li, Eric Xing, et~al. 2023.
\newblock Judging llm-as-a-judge with mt-bench and chatbot arena.
\newblock \emph{Advances in neural information processing systems}, 36:46595--46623.

\bibitem[{Zhou et~al.(2023)Zhou, Lu, Mishra, Brahma, Basu, Luan, Zhou, and Hou}]{zhou2023instruction}
Jeffrey Zhou, Tianjian Lu, Swaroop Mishra, Siddhartha Brahma, Sujoy Basu, Yi~Luan, Denny Zhou, and Le~Hou. 2023.
\newblock Instruction-following evaluation for large language models.
\newblock \emph{arXiv preprint arXiv:2311.07911}.

\end{thebibliography}
\bibliographystyle{acl_natbib}

\clearpage  

\appendix

\section{Dataset Details}
\label{appendix:data}

\begin{figure*}[t]
    \centering
    \includegraphics[width=1.0\textwidth]{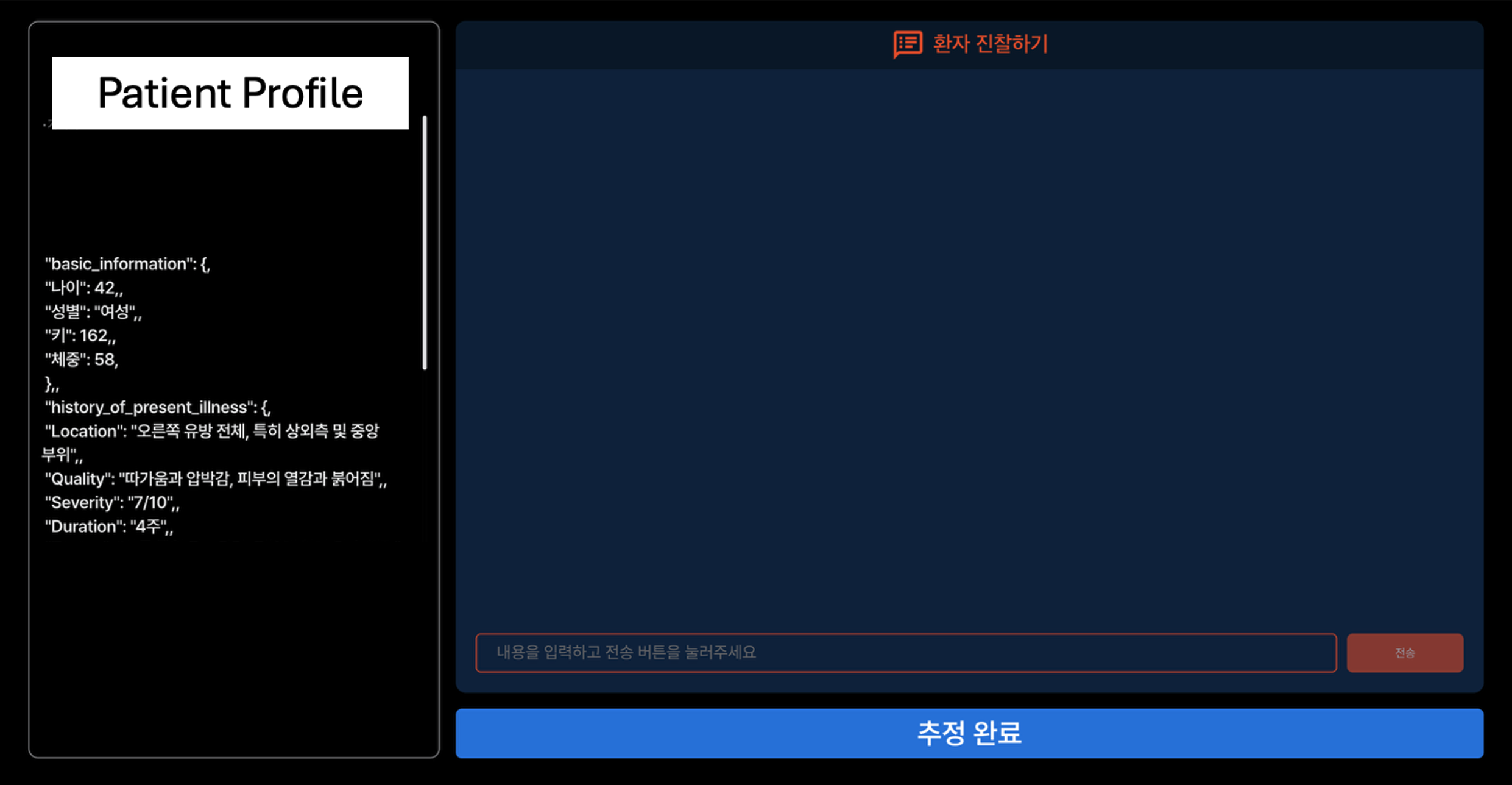}
    \caption{Interface of the medical pre-consultation platform where doctor and patient models engage in interactive pre-consultation dialogues.}
    \label{fig:platform}
\end{figure*}

\paragraph{Data Source and Scale}
Our real-world dataset was collected from a medical pre-consultation platform, where 40 licensed doctors conducted pre-consultations with patient models (see Figure~\ref{fig:platform}). In total, we collected 8,000 dialogues.

\paragraph{Real-World Distribution}
The dataset naturally exhibits a heavily skewed turn-count distribution, where most dialogues are short (see Table~\ref{tab:table4}). This inherent imbalance motivated our investigation into how turn-count distribution affects model behavior, especially in long medical dialogues.

\paragraph{Skewed Turn-Count Dataset}
To reflect the real-world data distribution, we constructed a Skewed Turn-Count Dataset by sampling 1,000 dialogues probabilistically from a pool of 8,000, preserving the original skewed turn-count distribution.

\paragraph{Uniform Turn-Count Dataset}
To address the Format Inertia phenomenon caused by the skewed turn-count distribution, we constructed a Uniform Turn-Count Dataset that equalizes the representation of dialogues across different lengths. The dataset creation procedure is outlined below:

\begin{enumerate}
\item \textbf{Turn-Based Binning:}
We first divided all 8,000 dialogues from the original dataset into 12 bins according to their maximum turn-count (from 1 to 12).

\item \textbf{Quota Determination:}  
To ensure uniformity, we used the bin with the smallest number of dialogues as the reference for sampling. As shown in Table~\ref{tab:table4}, the bin for 12-turn dialogues contains only 111 examples, which we set as the sampling quota.

\item \textbf{Uniform Sampling:}  
We then randomly sampled 111 dialogues without replacement from each of the nine bins, ranging from 4 to 12 turns (excluding bins with fewer than 4 turns, which contained no dialogues).

\item \textbf{Dataset Assembly:}  
The sampled dialogues were merged to form the final Uniform Turn-Count Dataset, consisting of 999 dialogues (111 dialogues $\times$ 9 bins), i.e., almost 1,000 in total. This construction ensures that the model is trained on a balanced distribution of short and long multi-turn interactions.
\end{enumerate}

\begin{table}[t]
\centering
\resizebox{0.3\columnwidth}{!}{%

\begin{tabular}{cc}
\toprule
\textbf{Turn} & \textbf{Count} \\
\midrule
1  & 0 \\
2  & 0 \\
3  & 0 \\
4  & 1835 \\
5  & 1890 \\
6  & 1825 \\
7  & 820 \\
8  & 705 \\
9  & 395 \\
10 & 260 \\
11 & 165 \\
12 & 111 \\
\bottomrule
\end{tabular}
}
\caption{Real-World Turn-Count Distribution in Medical Pre-Consultation.}
\label{tab:table4}
\end{table}

\section{Training Details}
\label{appendix:experiments}

\paragraph{Models}
We employed a set of Large Language Models (LLMs) as doctor models in our experiments.
Specifically, we used the instruction-tuned versions of the following models:
google/gemma-3-4b-it, 
google/gemma-3-12b-it, 
google/gemma-3-27b-it, and 
Qwen/Qwen2.5-3B-Instruct.

\paragraph{Fine-Tuning Procedure}
Supervised Fine-Tuning (SFT) was performed using the training datasets described in Appendix~\ref{appendix:data}. We utilized the Unsloth framework\footnote{\url{https://github.com/unslothai/unsloth}} to streamline the fine-tuning process. 

\paragraph{Hyperparameters}
We fine-tuned the models with the following hyperparameter configuration: a learning rate of $1\times10^{-4}$, batch size of 32, and 3 training epochs. We used QLoRA~\cite{dettmers2023qlora} with a rank of 32, lora alpha = 64, lora dropout = 0.1, a cosine learning rate scheduler, and a weight decay of 0.005.

\paragraph{Hardware Environment}
All fine-tuning experiments were conducted on NVIDIA A100 GPUs, each equipped with 80GB of memory.

\section{Patient Information Provided to Models}
\label{appendix:patient_info}

During evaluation, different sets of patient information were provided to the doctor and patient models to simulate realistic consultation dynamics.

\paragraph{Doctor Model Input}
As illustrated in Table~\ref{tab:table5}, the doctor model received a concise summary of the patient's profile. This summary includes essential information such as age, gender, chief complaint, and a brief description of symptoms. This limited view encourages the doctor model to ask clarifying questions to gather more detailed clinical information.

\paragraph{Patient Model Input}
Table~\ref{tab:table6} shows the full patient context that was provided to the patient model. This includes comprehensive clinical data such as detailed symptom descriptions, medical history, and other contextual information. The patient model responds to the doctor’s questions based on this full input, ensuring informative and contextually accurate answers.

This asymmetric information setup mimics real-world patient-doctor interactions and helps evaluate the models' ability to engage in effective diagnostic dialogue.

\section{System Prompts}
\label{appendix:prompts}

\subsection{Doctor System Prompt}
The following prompt was used for the doctor model to generate medically relevant follow-up questions:

\begin{promptbox}[verbatim]{Doctor System Prompt}
You are a physician with professional medical knowledge. Your task is to generate the optimal follow-up question that helps with differential diagnosis based on the given patient information and previous consultation history. \\

\#\#\# Instructions  \\
1. You must use only Korean. \\
2. Generate only questions that effectively collect medically useful information required for diagnosis. \\
3. Provide 2 to 5 options. \\
4. Do not include "Other" as an option. \\
5. Output in YAML format. \\
6. The question must end with "요?" (appropriate polite phrase in Korean). \\

\#\#\# Output format \\
Question:  \\
Options:  \\
... \\
\end{promptbox}

\subsection{Patient System Prompt}
The patient model received the following prompt along with specific patient profile information:

\begin{promptbox}[verbatim]{Patient System Prompt}
You are a patient with the following profile: \\
\{patient\_information\} \\

You should faithfully answer the doctor's inquiries for an appropriate diagnosis. Choose one of the questions provided by the doctor and respond. \\

Output format: \\
Answer: \\
\end{promptbox}

\section{Evaluation Details}
\label{appendix:evaluation}

\begin{figure*}[t]
    \centering
    \includegraphics[width=1.0\linewidth]{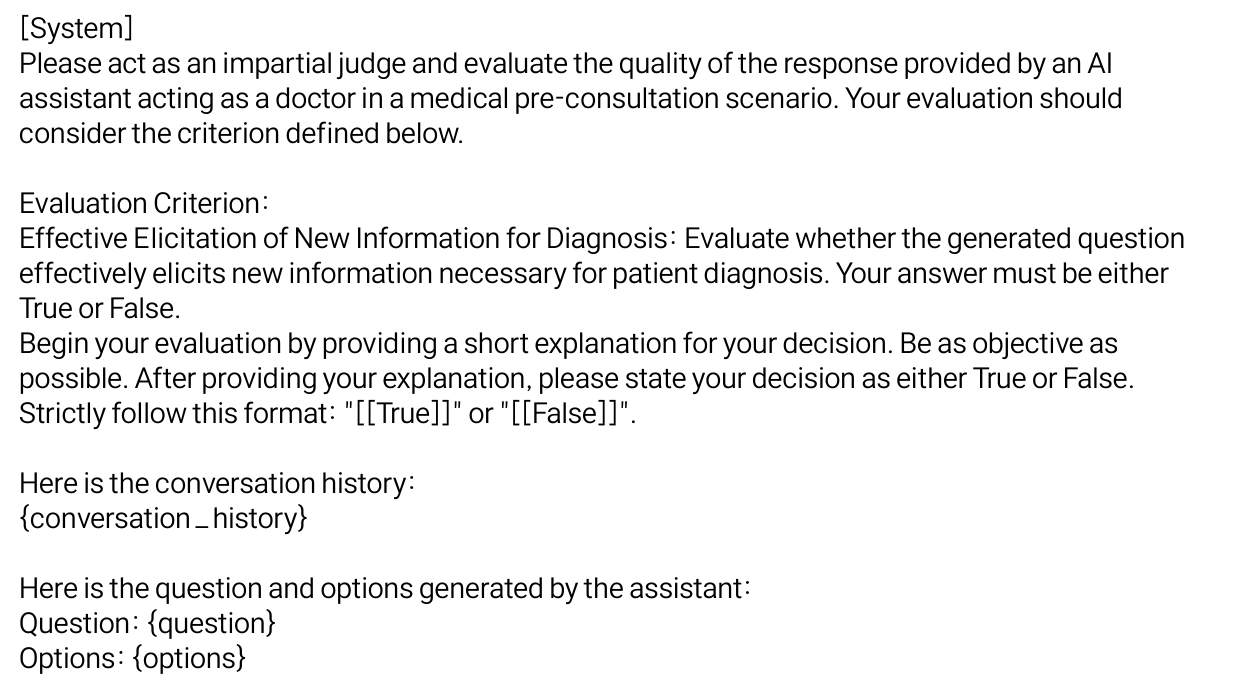}
    \caption{The Evaluation Prompt.}
    \label{fig:eval_prompt}
\end{figure*}

\paragraph{Judge LLM}
We employed GPT-4.1-mini~\footnote{gpt-4.1-mini-2025-04-14} as the evaluation judge. The model was prompted with a rubric to assess each turn based on format and task satisfaction.
We selected this model based on its strong performance on Korean medical QA benchmarks, particularly KorMedMCQA~\cite{kweon2024kormedmcqa}, which served as a key reference for evaluating its judgment capabilities.
The evaluation prompt used for the judge model is shown in Figure~\ref{fig:eval_prompt}.

\begin{table*}[t]
\centering
\resizebox{0.7\textwidth}{!}{%
\begin{tabular}{ll}
    \toprule
    Age & 45 \\
    Gender & Male \\
    Chief Complaint & Persistent cough and unintended weight loss \\
    Symptom Duration & Within 3 months \\
    Symptom Location & Right pectoral region and sternal area \\
    \bottomrule
\end{tabular}
}
\caption{Example of information provided to the doctor model during evaluation, representing a simplified summary of the patient's profile.}
\label{tab:table5}
\end{table*}

\begin{table*}[t]
\centering
\resizebox{0.7\textwidth}{!}{%
\begin{tabular}{ll}
    \toprule
    Disease & Pulmonary tuberculosis \\
    Department & Undetermined \\
    Typicality & Typical \\
    \midrule
    Age & 45 \\
    Gender & Male \\
    Height & 172 cm \\
    Weight & 68 kg \\
    \midrule
    Symptom Location & Central chest and occipital region \\
    Symptom Quality & Intermittent dry cough and chest tightness \\
    Symptom Severity & 5/10 \\
    Symptom Duration & 2 months \\
    Timing & Severe coughing in the morning, intermittent during day \\
    Context & Worsened after outdoor work and fatigue \\
    Modifying Factors & Warm tea and rest help; worsens with activity \\
    Associated Symptoms & Weight loss, night sweats, low-grade fever, mild dyspnea \\
    \midrule
    Pain Area & Right chest (pectoral), Sternal region \\
    Past History & No prior TB or chronic respiratory illness \\
    Social History & Construction worker, past smoker, high-density living \\
    Additional Info & Dust exposure, smoking history, persistent fatigue \\
    \bottomrule
\end{tabular}
}
\caption{Example of information accessible to the patient model during evaluation, including comprehensive clinical and contextual details.}
\label{tab:table6}
\end{table*}

\section{Constraint Examples}
\label{appendix:constraint_examples}
The examples provided are based on actual data used in our evaluation or similar scenarios.

\paragraph{Format Constraint Examples}
Figures~\ref{fig:ex_response_format}--\ref{fig:ex_sentence_startend} illustrate how a model's response can satisfy or violate each format constraint. Each figure shows a side-by-side comparison within a hypothetical dialogue turn where the model acts as a doctor.

\begin{figure*}[t]
\small
\hrule
\vspace{3mm}
\textbf{Constraint: \texttt{response\_format}} -- Whether the model's response adheres to the specified output format.
\vspace{3mm}
\hrule
\begin{minipage}[t]{0.48\textwidth}
\vspace{3mm}
\textbf{Satisfying Example}
\begin{verbatim}
question: "Where is the pain located?"
options:
  - "Entire head"
  - "Forehead"
  - "Back of the neck"
  - "Temples"
\end{verbatim}
\textbf{Explanation:} The response correctly follows the YAML structure.
\end{minipage}
\hfill
\begin{minipage}[t]{0.48\textwidth}
\vspace{3mm}
\textbf{Violating Example}
\begin{verbatim}
Where is the pain? The options are 1. Entire head,
2. Forehead, 3. Back of the neck, 4. Temples.
\end{verbatim}
\textbf{Explanation:} The response does not follow the specified YAML format and consists only of plain text.

\end{minipage}
\vspace{3mm}
\hrule
\vspace{3mm}
\caption{Examples for the \texttt{response\_format} constraint.}
\label{fig:ex_response_format}
\end{figure*}

\begin{figure*}[t]
\small
\hrule
\vspace{3mm}
\textbf{Constraint: \texttt{response\_language}} -- Whether the model's response is composed exclusively in the specified language (e.g., English).
\vspace{3mm}
\hrule
\begin{minipage}[t]{0.48\textwidth}
\vspace{3mm}
\textbf{Satisfying Example}
\begin{verbatim}
question: "When did the symptoms start?"
options:
  - "Today"
  - "Yesterday"
  - "A few days ago"
  - "More than a week ago"
\end{verbatim}
\textbf{Explanation:} All content in the response is written in English.
\vspace{3mm}
\end{minipage}
\hfill
\begin{minipage}[t]{0.48\textwidth}
\vspace{3mm}
\textbf{Violating Example}
\begin{verbatim}
question: "Quand les symptômes ont-ils commencé ?"
options:
  - "Aujourd'hui"
  - "Hier"
\end{verbatim}
\textbf{Explanation:} The response is written in French, violating the English-only requirement.
\end{minipage}
\hrule
\vspace{3mm}
\caption{Examples for the \texttt{response\_language} constraint.}
\label{fig:ex_response_language}
\end{figure*}

\begin{figure*}[t]
\small
\hrule
\vspace{3mm}
\textbf{Constraint: \texttt{forbidden\_words}} -- Whether the response uses any forbidden words specified by the service (e.g., a forbidden word: \texttt{"Other"}).
\vspace{3mm}
\hrule
\begin{minipage}[t]{0.48\textwidth}
\vspace{3mm}
\textbf{Satisfying Example}
\begin{verbatim}
question: "What type of painkiller did you take?"
options:
  - "Ibuprofen"
  - "Acetaminophen"
  - "Aspirin"
  - "None"
\end{verbatim}
\textbf{Explanation:} The model avoids the forbidden word "Other" by listing only specific options.
\vspace{3mm}
\end{minipage}
\hfill
\begin{minipage}[t]{0.48\textwidth}
\vspace{3mm}
\textbf{Violating Example}
\begin{verbatim}
question: "What type of painkiller did you take?"
options:
  - "Ibuprofen"
  - "Acetaminophen"
  - "Other"
  - "None"
\end{verbatim}
\textbf{Explanation:} The model includes the forbidden word "Other" in the options, violating the constraint.
\end{minipage}
\hrule
\vspace{3mm}
\caption{Examples for the \texttt{forbidden\_words} constraint where "Other" is a forbidden term.}
\label{fig:ex_forbidden_words}
\end{figure*}

\begin{figure*}[t]
\small
\hrule
\vspace{3mm}
\textbf{Constraint: \texttt{number\_options}} -- Whether the number of options provided with the question is correct (e.g., a rule: must be between 2 and 5 options).
\vspace{3mm}
\hrule
\begin{minipage}[t]{0.48\textwidth}
\vspace{3mm}
\textbf{Satisfying Example}
\begin{verbatim}
question: "What is the nature of the pain?"
options:
  - "Throbbing"
  - "Stabbing"
  - "Squeezing"
\end{verbatim}
\textbf{Explanation:} It provides 3 options, which adhere to the rule.
\end{minipage}
\hfill
\begin{minipage}[t]{0.48\textwidth}
\vspace{3mm}
\textbf{Violating Example}
\begin{verbatim}
question: "Where is the pain located?"
options:
  - "Forehead"
  - "Temples"
  - "Back of the head"
  - "Neck"
  - "Jaw"
  - "Behind the eyes"
  - "Left side"
  - "Right side"
\end{verbatim}
\textbf{Explanation:} It provides 8 options, exceeding the maximum allowed number.
\vspace{3mm}

\end{minipage}
\hrule
\vspace{3mm}
\caption{Examples for the \texttt{number\_options} constraint.}
\label{fig:ex_number_options}
\end{figure*}

\begin{figure*}[t]
\small
\hrule
\vspace{3mm}
\textbf{Constraint: \texttt{sentence\_startend}} -- Whether the question ends with the appropriate polite phrase (e.g., must end with "요?").
\vspace{3mm}
\hrule
\begin{minipage}[t]{0.48\textwidth}
\vspace{3mm}
\textbf{Satisfying Example}
\begin{verbatim}
question: "통증이 가장 심한 시간대가 언제인가요?"
options:
  - "아침"
  - "오후"
  - "밤"
  - "특정 시간 없음"
\end{verbatim}
\textbf{Explanation:} The question ends with a polite expression ("요?"), conforming to the expected sentence structure.
\end{minipage}
\hfill
\begin{minipage}[t]{0.48\textwidth}
\vspace{3mm}
\textbf{Violating Example}
\begin{verbatim}
question: "통증이 가장 심한 시간대?"
options:
  - "아침"
  - "오후"
  - "밤"
  - "없음"
\end{verbatim}
\textbf{Explanation:} The question ends with a fragment, not a polite sentence ending with "요?", violating the constraint.
\vspace{3mm}

\end{minipage}
\hrule
\vspace{3mm}
\caption{Examples for the \texttt{sentence\_startend} constraint requiring questions to end with "요?".}
\label{fig:ex_sentence_startend}
\end{figure*}

\paragraph{Task Constraint Examples}
Figure~\ref{fig:ex_clinical_utility} provides an example for the clinical\_utility constraint. It shows whether the model's (doctor's) question effectively elicits new information necessary for patient diagnosis.


\section{Real-world Case Study}
Beyond our controlled experimental setting, we applied this model to VDoc\footnote{\url{https://www.vdoc.kr/en}}, a medical pre-consultation service deployed in South Korea, where we directly observed Format Inertia in real-world production settings. In one notable case, a patient consultation extended to 15 turns due to complex symptoms. After turn 8, the model began repeating variations of previously asked questions about symptom onset and pain location, despite having already received comprehensive answers. This real-world instance confirmed that Format Inertia is not merely an experimental artifact but a critical production issue that impacts user experience and diagnostic efficiency.

\section{Distinction Between Format Inertia and Context Rot}
While Format Inertia may appear related to the concept of ``Context Rot''---a general performance degradation over long contexts---we argue that Format Inertia represents a more specific and actionable failure mode:

\begin{itemize}
    \item \textbf{A Specious Failure Mode:} Context Rot describes broad degradation across all metrics. In contrast, Format Inertia identifies a deceptive scenario where models maintain high Format-Constraint Satisfaction Rate (FCSR) while experiencing sharp drops in Task-Constraint Satisfaction Rate (TCSR). The model appears functional but fails at its core diagnostic purpose.
    
    \item \textbf{Specific Pattern Recognition:} Format Inertia manifests as repetitive questioning patterns that are formally correct but diagnostically useless. This specific behavioral pattern makes it identifiable and addressable through targeted interventions like our uniform sampling approach.
    
    \item \textbf{Production Impact:} Unlike general context degradation, Format Inertia creates a particularly problematic user experience where patients receive seemingly valid but redundant questions, leading to frustration and abandonment of the consultation process.
\end{itemize}

\begin{figure*}[t]
\small
\hrule
\vspace{3mm}
\textbf{Constraint: \texttt{clinical\_utility}} -- Whether the generated question effectively elicits new information required for patient diagnosis.
\vspace{3mm}
\hrule
\begin{minipage}[t]{0.48\textwidth}
\vspace{3mm}
\textbf{Satisfying Example: (Dialogue Turn 5)}
\textbf{Previous conversation summary:} The patient has mentioned having a headache, located in the forehead, which started a few days ago.
\begin{verbatim}
question: "Were there any specific activities or 
  changes (e.g., stress, lack of sleep, dietary 
  changes) before the headache occurred?"
options:
  - "Yes, there were"
  - "No, there were not"
\end{verbatim}
\textbf{Explanation:} This is a question that explores new information about the background of the patient's symptoms, attempting to obtain crucial information for diagnosis. It asks about a new aspect not covered in the previous conversation.
\end{minipage}
\hfill
\begin{minipage}[t]{0.48\textwidth}
\vspace{3mm}
\textbf{Violating Example (Format Inertia Case): (Dialogue Turn 10)}
\textbf{Previous conversation summary:} The patient has already provided a lot of information, including the headache, its location, onset, nature of the pain, and accompanying symptoms (no nausea, sensitivity to light/sound). The model had also asked "When did you start feeling sick?" in turn 4.
\begin{verbatim}
question: "When did the symptoms start?"
options:
  - "Today"
  - "Yesterday"
  - "A few days ago"
  - "More than a week ago"
\end{verbatim}
\textbf{Explanation:} The model is asking a question that has already been answered or can be inferred from a previous turn. Although it is formally a question, it is an unproductive one that fails to elicit any new diagnostic information.
\vspace{3mm}

\end{minipage}
\hrule
\vspace{3mm}
\caption{Example for the \texttt{clinical\_utility} constraint.}
\label{fig:ex_clinical_utility}
\end{figure*}

\end{document}